\documentclass[conference]{IEEEtran}
\IEEEoverridecommandlockouts
\usepackage[T1]{fontenc}
\usepackage[utf8]{inputenc} % remove if using XeLaTeX/LuaLaTeX
\usepackage{lmodern}

\usepackage{microtype}     % better spacing, fewer over/underfull boxes
\usepackage{amsmath,amssymb,amsfonts}
\usepackage{siunitx}

\usepackage{graphicx}
\usepackage{array}
\usepackage{booktabs}
\usepackage{multirow,tabularx}
\usepackage{adjustbox}

\usepackage{xcolor}
\usepackage{textcomp}
\usepackage{enumitem}
\setlist[enumerate]{leftmargin=*, itemsep=1pt, topsep=2pt}
\usepackage[section]{placeins}

\usepackage{newunicodechar}
\newunicodechar{—}{\textemdash{}}
\newunicodechar{Δ}{\ensuremath{\Delta}}
\newunicodechar{∝}{\ensuremath{\propto}}

\PassOptionsToPackage{hyphens}{url}

\usepackage[unicode,breaklinks=true,colorlinks=true,linkcolor=blue,citecolor=blue,urlcolor=blue]{hyperref}

\usepackage{algorithmic}

\newcommand{\code}[1]{\texttt{#1}}      
\usepackage{booktabs}
\usepackage{siunitx}
\usepackage{multirow}
\usepackage{threeparttable}

\usepackage[T1]{fontenc}
\usepackage[utf8]{inputenc}

\DeclareRobustCommand{\code}[1]{\texttt{\detokenize{#1}}}   % underscores OK
\DeclareRobustCommand{\apiname}[1]{\code{#1}}               % alias for APIs

\newenvironment{purple}{
    \color{purple}
}{
    \ignorespacesafterend
}

\begin{document}

\title{Delta-Audit: Explaining What Changes When Models Change

\thanks{Code and reproducible artifacts: \mbox{\url{https://github.com/arshiahemmat/delta-audit}}}
}

%\author{\IEEEauthorblockN{1\textsuperscript{st} Arshia Hemmat}
\author{\IEEEauthorblockN{1\textsuperscript{st} Arshia Hemmat}
\IEEEauthorblockA{\textit{dept. Computer Engineering} \\
\textit{University of Isfahan}\\
Isfahan, Iran \\
arshiahemmat@mehr.ui.ac.ir}
\and
\
\IEEEauthorblockN{2\textsuperscript{nd} Afsaneh Fatemi}
\IEEEauthorblockA{\textit{dept. Computer Engineering} \\
\textit{University of Isfahan}\\
Isfahan, Iran \\
a\_fatemi@eng.ui.ac.ir}
}

\maketitle

\begin{abstract}
\label{sec:abstract}
Model updates (new hyperparameters, kernels, depths, solvers, or data) change performance, but the \emph{reason} often remains opaque. We introduce \textbf{Delta-Attribution} (\mbox{$\Delta$-Attribution}), a model-agnostic framework that explains \emph{what changed} between versions $A$ and $B$ by differencing per-feature attributions: $\Delta\phi(x)=\phi_B(x)-\phi_A(x)$. We evaluate $\Delta\phi$ with a \emph{$\Delta$-Attribution Quality Suite} covering magnitude/sparsity (L1, Top-$k$, entropy), agreement/shift (rank-overlap@10, Jensen--Shannon divergence), behavioural alignment (Delta Conservation Error, DCE; Behaviour--Attribution Coupling, BAC; CO$\Delta$F), and robustness (noise, baseline sensitivity, grouped occlusion). 

Instantiated via fast occlusion/clamping in standardized space with a class-anchored margin and baseline averaging, we audit 45 settings: five classical families (Logistic Regression, SVC, Random Forests, Gradient Boosting, $k$NN), three datasets (Breast Cancer, Wine, Digits), and three A/B pairs per family. \textbf{Findings.} Inductive-bias changes yield large, behaviour-aligned deltas (e.g., SVC poly$\!\rightarrow$rbf on Breast Cancer: BAC$\approx$0.998, DCE$\approx$6.6; Random Forest feature-rule swap on Digits: BAC$\approx$0.997, DCE$\approx$7.5), while ``cosmetic'' tweaks (SVC \texttt{gamma=scale} vs.\ \texttt{auto}, $k$NN search) show rank-overlap@10$=1.0$ and DCE$\approx$0. The largest redistribution appears for deeper GB on Breast Cancer (JSD$\approx$0.357). $\Delta$-Attribution offers a lightweight update audit that complements accuracy by distinguishing benign changes from behaviourally meaningful or risky reliance shifts.

\end{abstract}


\section{Introduction}
\label{sec:intro}
Models rarely stand still. In modern ML systems, practitioners routinely update models by changing hyperparameters, switching architectures, fine-tuning on fresh data, or compressing for deployment. These updates can shift performance in obvious ways (accuracy goes up or down), yet the \emph{reason} for the shift often remains opaque: \emph{what} parts of the decision logic changed, \emph{where} did the model start relying more (or less) on particular features, and \emph{do} those changes align with observed behaviour?

Most explanation methods answer a different question: they explain \emph{one} model at a time---e.g., via additive attributions such as SHAP~\cite{Lundberg2017SHAP,Lundberg2020TreeSHAP}, rule-based local explanations such as Anchors~\cite{Ribeiro2018Anchors}, or perturbation/occlusion maps~\cite{Zeiler2014}. A sizeable body of work highlights stability concerns for such explanations~\cite{Adebayo2018Sanity,Slack2020Fooling}, and recent papers begin to study monitoring of explanations or ``explanation shift'' under distribution shift~\cite{Mougan2022ExplShift,Mougan2025ExplShift}. However, in practical pipelines the more immediate need is often an \emph{update audit}: when we replace model $A$ with model $B$, how did the model’s reliance on input features change, and is that change consistent with the observed change in outputs?

\textbf{Problem.} We study this update-audit question for supervised classification. Given any attribution method that produces per-feature scores $\phi_A(x)$ and $\phi_B(x)$ for models $A$ and $B$, we define the \emph{delta attribution}
\[
\Delta\phi(x) \;=\; \phi_B(x) - \phi_A(x),
\]
and we evaluate the quality of $\Delta\phi$ with respect to: (i) \emph{magnitude and concentration} (are changes small and diffuse or large and focused?), (ii) \emph{agreement and distributional shift} between the two explanation vectors, (iii) \emph{behavioural alignment} with the observed output change $\Delta f(x)$, and (iv) \emph{robustness} to noise and baseline choices. Intuitively, $\Delta\phi$ should highlight where the new model reallocated reliance; a good $\Delta$ explanation should \emph{co-move} with behaviour changes and remain stable to small input perturbations or reasonable baseline choices.

\textbf{Approach (Δ-Attribution).} We propose \textbf{Delta-Attribution} (Δ-Attribution), a simple, model-agnostic framework that turns any local explainer into an \emph{update explainer}. In this paper we instantiate it with a fast \emph{occlusion/clamping} explainer in standardized feature space: for feature $j$, we set $x_j$ to a baseline and measure the margin drop for a chosen class; the attribution is the drop, and $\Delta\phi$ is the difference between models. To connect explanations to behaviour, we define $f(x)$ as the class-specific margin for a fixed reference class (in our runs, the class predicted by $B$), yielding $\Delta f(x) = f_B(x) - f_A(x)$. We then compute a \emph{Δ-Attribution Quality Suite} comprising:
\begin{itemize}
  \item \underline{Internal Δ metrics}: $\Delta$ magnitude ($\ell_1$), Top-$K$ concentration, entropy, rank overlap@10, and Jensen–Shannon divergence between normalized $|\phi_A|$ and $|\phi_B|$.
  \item \underline{Behaviour-linked Δ metrics}: the \emph{Delta Conservation Error} (DCE) $=\; \mathbb{E}_x\,\big| \sum_j \Delta\phi_j(x) - \Delta f(x)\big|$, the \emph{Behaviour–Attribution Coupling} (BAC; Pearson corr of $\|\Delta\phi\|_1$ with $|\Delta f|$), and class-outcome focus (COΔF) that checks whether $\Delta$ mass concentrates on features deemed globally relevant by the updated model when fixes/regressions occur.
  \item \underline{Robustness}: sensitivity of $\Delta\phi$ to Gaussian input noise and to alternative baselines (mean vs.\ median), plus a grouped-occlusion stress-test that jointly clamps top-$k$ features to capture interactions.
\end{itemize}

\textbf{Why not single-model explanations?} Single-model inspections can show \emph{what} a model currently relies on, but they leave the \emph{update} unanswered. Directly differencing attributions provides a concrete, low-friction view of what changed. Importantly, our quality suite guards against over-interpretation: e.g., high DCE warns that a purely additive occlusion view may be unreliable; low rank overlap with high JSD indicates a genuine redistribution rather than mere reweighting; and robustness checks catch baseline- or noise-fragile deltas.

\textbf{Scope and setting.} We aim for a practical tool that is cheap enough to run during everyday model iteration. We therefore avoid heavy model-specific explanation tooling and large-scale hyperparameter sweeps. Our study intentionally targets \emph{classical} ML families (logistic regression, SVM, random forests, gradient boosting, $k$NN) across three standard tabular/image datasets (Breast Cancer, Wine, Digits). For each family we construct three A/B pairs that toggle inductive bias (e.g., kernel, depth) or regularization/solver choices, so that we can observe small vs.\ large update regimes within the same learner.

\textbf{Research questions.} We organize the study around three questions:
\begin{description}
  \item[RQ1 — Internal change:] How do $\Delta$ magnitude, sparsity/concentration, and rank agreement behave across small vs.\ large updates within and across algorithms?
  \item[RQ2 — Behavioural alignment:] When performance changes, does $\|\Delta\phi\|_1$ increase and does DCE decrease? Do COΔF scores indicate that fixes concentrate $\Delta$ mass on globally relevant features for $B$?
  \item[RQ3 — Robustness:] Are the observed $\Delta$ patterns stable under input noise and alternative baselines, and do grouped occlusions substantially alter conclusions (indicative of interactions)?
\end{description}

\textbf{Contributions.}
\begin{itemize}
  \item We formalize \emph{Delta-Attribution} as a model-agnostic lens for \emph{explaining updates}: $\Delta\phi(x)=\phi_B(x)-\phi_A(x)$, with a quality suite that measures magnitude/sparsity, agreement/shift, behavioural alignment (DCE, BAC, COΔF), and robustness.
  \item We provide an efficient instantiation via standardized occlusion with baseline averaging (mean/median) and a grouped-occlusion stress-test, together with implementation notes (class-anchored margins, reproducible pipelines).
  \item We run a broad empirical audit covering five ML families, three datasets, and nine A/B pairs per family. We show that changes in \emph{inductive bias} (e.g., kernel or depth) produce large, behaviour-aligned $\Delta$, whereas ``cosmetic'' knobs (e.g., SVC $\gamma{=}$scale vs.\ auto) yield tiny, concentrated $\Delta$ and near-perfect rank overlap.
  \item We release \emph{Delta-Attribution} as a lightweight platform: scripts, metrics, and publication-ready assets for plug-in updates and future benchmarks.
\end{itemize}

\textbf{Positioning.} Our focus complements single-model explainability~\cite{Lundberg2017SHAP,Ribeiro2018Anchors,Zeiler2014} and explanation-stability work~\cite{Adebayo2018Sanity,Slack2020Fooling,Chen2024R2ET}. Unlike distribution-shift attribution~\cite{Federici2021InfoShift,Zhang2023WhyFail,Mougan2022ExplShift,Mougan2025ExplShift}, we target \emph{version-to-version} changes under fixed test distributions, mirroring the common operational reality of model updates. As rapid editing and fine-tuning become routine (e.g., model editing~\cite{Meng2022ROME,Mitchell2022MEND,Meng2023MEMIT,Gupta2024SeqEdit}), update-centric explanations provide a governance signal that complements accuracy and fairness dashboards.

\textbf{Takeaway.} Δ-Attribution turns existing local explainers into a practical tool for update audits. By quantifying \emph{how} reliance shifts and whether those shifts \emph{explain} behaviour deltas, our framework helps practitioners decide when an update is benign, when it meaningfully improves reliance on task-relevant signals, and when it warrants further scrutiny.

\section{Related Work}
\label{sec:related}
\textbf{Post-hoc explanations for a single model.}
Feature-attribution and exemplar methods explain individual model decisions via local scores or rules.
Canonical approaches include LIME and Anchors~\cite{Ribeiro2018Anchors}, SHAP~\cite{Lundberg2017SHAP} and its tree-exact variant TreeSHAP~\cite{Lundberg2020TreeSHAP}, Integrated Gradients~\cite{Sundararajan2017IG}, SmoothGrad~\cite{Smilkov2017SmoothGrad}, gradient-based visualizations such as Grad-CAM~\cite{Selvaraju2020GradCAM}, and occlusion/clamping~\cite{Zeiler2014}.
While these tools are widely used, several works highlight pitfalls:
saliency “sanity checks’’ show some maps ignore model or data~\cite{Adebayo2018Sanity};
ROAR finds many methods fail to remove truly important evidence upon retraining~\cite{Hooker2019ROAR};
and input-perturbation methods can be manipulated adversarially~\cite{Slack2020Fooling}.
These observations motivate explanation diagnostics beyond visual appeal. :contentReference[oaicite:0]{index=0}

\textbf{Stability, monitoring, and evaluation.}
Beyond point explanations, recent work proposes to \emph{measure} explanation reliability and track it over time.
R2ET trains for stable top-$k$ saliency at little extra cost~\cite{Chen2024R2ET}.
Explanation-shift research argues that monitoring \emph{changes} in explanation distributions can be a more sensitive indicator of behavior change than monitoring input/output distributions alone, with formal detectors on tabular data~\cite{Mougan2022ExplShift,Mougan2025ExplShift}.
Complementary studies evaluate robustness of attribution maps under perturbations and propose stronger evaluation protocols~\cite{Nieradzik2024Reliable, Wang2020SmoothedGeom}.
Our work contributes here by defining a \emph{version-to-version} suite (Δ-Attribution Quality Suite) that quantitatively captures magnitude, sparsity, rank agreement, distributional shift (JSD), behavior linkage (DCE, BAC, COΔF), and robustness (noise and grouped occlusion). :contentReference[oaicite:1]{index=1}

\textbf{Explaining performance changes under distribution shift.}
A parallel line attributes \emph{error changes} across environments.
Federici et al.\ give an information-theoretic account of shift sources and error decompositions~\cite{Federici2021InfoShift};
Zhang et al.\ attribute performance deltas to causal shift factors using Shapley-style games~\cite{Zhang2023WhyFail}.
These focus on \emph{what caused} performance change across datasets.
In contrast, Δ-Attribution inspects \emph{how a model’s reliance redistributes over features} when the model itself changes (fine-tuning, hyperparameters, editing), and links those reliance deltas to behavior deltas via DCE/BAC/COΔF. :contentReference[oaicite:2]{index=2}

\textbf{Model editing and rapid updates.}
Frequent post-deployment updates—ROME~\cite{Meng2022ROME}, MEND~\cite{Mitchell2022MEND}, MEMIT~\cite{Meng2023MEMIT}, and stability fixes for sequential editing~\cite{Gupta2024SeqEdit}—make version-aware explanations increasingly important.
Our Δ-Attribution provides a lightweight audit for such updates, orthogonal to the editing method itself. :contentReference[oaicite:3]{index=3}

\section{Proposed Approach}
\label{sec:approach}
\subsection{Setup and notation}
Let $f_A,f_B$ be two model versions on $x\in\mathbb{R}^d$.
An explainer $E$ returns per-feature scores $\phi_f(x)\in\mathbb{R}^d$.
We study the delta attribution
\[
\Delta\phi(x)=\phi_B(x)-\phi_A(x),
\]
and assess its quality with a dedicated metric suite.

\paragraph{Reference class and score.}
For each $x$ we fix the reference class $c(x)$ as the class predicted by $f_B$.
We score that class using the model’s margin or log-odds:
\[
f(x)=
\begin{cases}
[\mathrm{dec}(x)]_{c(x)}, & \text{if function is available},\\
\log\!\dfrac{p_{c(x)}(x)}{1-p_{c(x)}(x)}, & \text{if probabilities are available},\\
\log\!\dfrac{p_{c(x)}(x)+\varepsilon}{1-p_{c(x)}(x)+\varepsilon},\! & \text{otherwise},\ \varepsilon=10^{-9}.
\end{cases}
\]
Here $\mathrm{dec}(\cdot)$ denotes the model's decision function
(\apiname{decision_function}); $p(\cdot)$ comes from \apiname{predict_proba}.
Anchoring to $c(x)$ compares the same class across versions and stabilizes
behaviour-linked metrics (BAC, DCE).

\subsection{Explainer: occlusion/clamping}
We use a fast, model-agnostic occlusion explainer~\cite{Zeiler2014} in standardized space.
Let $b$ be a training-set baseline (mean by default; optionally averaged with the median).
For feature $j$,
\[
\phi_{f,j}(x)=f(x)-f(x_{-j}),\qquad
x_{-j}:\ x_j\leftarrow b_j .
\]
We compute $\phi_A,\phi_B$ with the \emph{same} baseline and the same $c(x)$.

\paragraph{Grouped-occlusion stress test.}
To probe interactions, jointly clamp the top-$k$ features (by $|\phi_B|$), recompute $\Delta\phi$, and report
\[
\rho=\mathbb{E}\Big[\frac{\|\Delta\phi(x)\|_1}{\|\Delta\phi^{(\text{group-}k)}(x)\|_1+\epsilon}\Big].
\]
Large $\rho$ indicates a few features drive the change.

\subsection{The $\Delta$-Attribution Quality Suite}
Averages are over the test set (or a stratified subset of up to 256 samples).

\paragraph{\textbf{Internal $\Delta$ metrics.}}
Let $u(x)=|\Delta\phi(x)|$ and $s(x)=u(x)/\|u(x)\|_1$ (skip if $\|u\|_1=0$).
\begin{enumerate}[leftmargin=1.2em,itemsep=2pt,topsep=2pt]
\item {Magnitude:} $\mathbb{E}\,\|\Delta\phi(x)\|_1$.
\item {Concentration:} $\Delta\mathrm{TopK}@10=\mathbb{E}\!\left[\sum_{j\in\mathrm{Top10}(u)} s_j(x)\right]$;
      {Entropy:} $\mathbb{E}\!\left[-\sum_j s_j(x)\log s_j(x)\right]$.
\item {Rank agreement:} Jaccard overlap of $\mathrm{Top10}(|\phi_A|)$ vs.\ $\mathrm{Top10}(|\phi_B|)$ (mean/median).
\item {Distributional shift:} $\mathbb{E}\big[\mathrm{JSD}(p\Vert q)\big]$ with
      $p=|\phi_A|/\||\phi_A|\|_1$, $q=|\phi_B|/\||\phi_B|\|_1$~\cite{Lin1991JSD}.
\end{enumerate}

\paragraph{\textbf{Behaviour-linked $\Delta$ metrics.}}
Let $\Delta f(x)=f_B(x)-f_A(x)$.
\begin{enumerate}[leftmargin=1.2em,itemsep=2pt,topsep=2pt]
\item {DCE:} $\mathbb{E}\big|\sum_j\Delta\phi_j(x)-\Delta f(x)\big|$ (diagnostic; smaller is better).
\item {BAC:} $\mathrm{corr}_x\!\big(\|\Delta\phi(x)\|_1,\;|\Delta f(x)|\big)$.
\item {CO$\Delta$F:} using the top-$m$ relevant features for $f_B$ from permutation importance ($m{=}10$), report the fraction of $\Delta$ mass on that set for \emph{fixes} and for \emph{regressions}.
\end{enumerate}

\paragraph{\textbf{Robustness.}}
\begin{enumerate}[leftmargin=1.2em,itemsep=2pt,topsep=2pt]
\item {$\Delta$-stability:} for $\varepsilon\!\sim\!\mathcal{N}(0,\sigma^2 I)$ with $\sigma\!\in\!\{0.01,0.05\}$,
$\mathbb{E}\,\|\Delta\phi(x+\varepsilon)-\Delta\phi(x)\|_1/(\|\varepsilon\|_2+\epsilon)$.
\item {Grouped occlusion ratio:} $\rho$ above.
\end{enumerate}

\paragraph{Note on additivity.}
Path methods such as Integrated Gradients~\cite{Sundararajan2017IG} satisfy additivity; occlusion does not.
Hence DCE is a \emph{diagnostic} rather than expected to be zero.

\section{EXPERIMENTAL SETUP}
\label{sec:experiments}
\begin{figure*}[t]
  \centering
  \includegraphics[width=\linewidth]{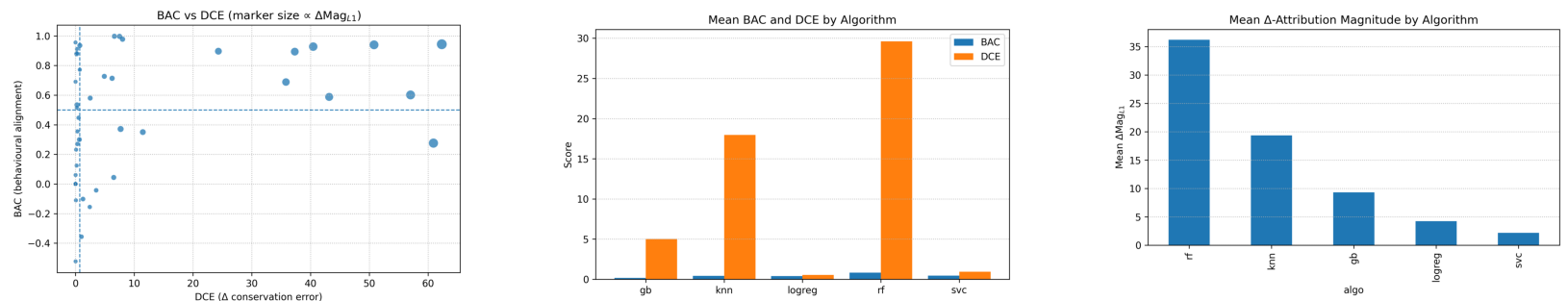}
  \caption{Overview of $\Delta$-Attribution results across all (dataset, algorithm, pair).
  (a) BAC vs.\ DCE with marker size $\propto \|\Delta\phi\|_1$;
  (b) mean BAC and DCE by algorithm; (c) mean $\Delta$-magnitude by algorithm.}
  \label{fig:delta_overview}
\end{figure*}

\subsection{Datasets and Preprocessing}
We use three standard \texttt{scikit-learn} datasets:
Breast Cancer (binary; $n{=}569$, $d{=}30$), Wine (3-class; $n{=}178$, $d{=}13$), and Digits (10-class; $n{=}1797$, $8{\times}8$ images flattened to $d{=}64$).
Each dataset is split with a stratified 80/20 train--test split (\texttt{random\_state}$=42$).
A \texttt{StandardScaler} is fit on training data and applied to test data.
All models are wrapped in a \texttt{Pipeline} so that versions $A$ and $B$ share identical preprocessing.

\subsection{Learners and A/B Configurations}
We study five families: logistic regression (\apiname{logreg}), support vector machines
(\apiname{svc} with \apiname{probability=true}), random forests (\apiname{rf}),
gradient boosting (\apiname{gb}), and $k$-nearest neighbors (\apiname{knn}). Each family
has three A/B pairs that toggle regularization, inductive bias (kernel/depth), or
search strategy (Table~\ref{tab:ab_pairs}). All models are wrapped in a
\apiname{Pipeline} with a shared \apiname{StandardScaler}.

\subsection{Score Function and Explainer}
For each test sample we anchor to the class predicted by $f_B$. We score that class
using the model margin when available (\apiname{decision_function}); otherwise we use
log-odds from \apiname{predict_proba}. Attributions are computed with occlusion/clamping
in standardized space: for feature $j$, clamp $x_j$ to a shared training baseline $b_j$
and define
\[
\phi_{f,j}(x)=f(x)-f(x_{-j}) \quad\text{with}\quad x_{-j}:\ x_j \leftarrow b_j.
\]
To reduce baseline artifacts we average mean/median baselines when both are available.
We always use the same baseline and the same reference class for $A$ and $B$.

\subsection{The $\Delta$-Attribution Suite}
Let $\Delta\phi(x)=\phi_B(x)-\phi_A(x)$, $u(x)=|\Delta\phi(x)|$, and $s(x)=u(x)/\|u(x)\|_1$ (skip samples with $\|u\|_1=0$).  
We report the metrics in Table~\ref{tab:delta_glossary}. Briefly:
(i) \textit{Magnitude} $\mathbb{E}\|\Delta\phi\|_1$;  
(ii) \textit{Concentration} $\Delta\mathrm{TopK}@10=\mathbb{E}[\sum_{j\in \mathrm{Top10}(u)} s_j]$ and entropy $\mathbb{E}[-\sum_j s_j\log s_j]$;  
(iii) \textit{Rank agreement} (Jaccard overlap of Top-10$(|\phi_A|)$ vs.\ Top-10$(|\phi_B|)$);  
(iv) \textit{Distributional shift} $\mathrm{JSD}(|\phi_A|,|\phi_B|)$~\cite{Lin1991JSD};  
(v) \textit{Behaviour linkage} with $\Delta f=f_B-f_A$: DCE $=\mathbb{E}\big|\sum_j\Delta\phi_j-\Delta f\big|$ and BAC $=\mathrm{corr}(\|\Delta\phi\|_1,|\Delta f|)$;  
(vi) \textit{Robustness}: $\Delta$-stability to Gaussian noise ($\sigma\in\{0.01,0.05\}$) and a grouped-occlusion ratio (jointly clamping top-$k{=}2$ features).

\textbf{Assets used in this paper.}
We aggregate the $\Delta$-suite by algorithm (Table~\ref{tab:delta_summary_by_algo}), plot a three-panel overview (Fig.~\ref{fig:delta_overview}), and list per-dataset Top-5 A/B pairs by $\Delta$ magnitude (Table~\ref{tab:top5_deltamag}).

\begin{table}[t]
\centering
\caption{A/B configurations per learner family (all models in a shared \texttt{Pipeline}).}
\label{tab:ab_pairs}
\setlength{\tabcolsep}{3.5pt}
\renewcommand{\arraystretch}{1.03}
\begin{tabular}{l l l}
\toprule
\textbf{Family} & \textbf{Pair A} & \textbf{Pair B} \\
\midrule
\texttt{logreg}\,P1 & C=1.0 (l2, lbfgs) & C=0.1 (l2, lbfgs) \\
\texttt{logreg}\,P2 & l2 (liblinear, C=1.0) & l1 (liblinear, C=1.0) \\
\texttt{logreg}\,P3 & lbfgs (l2, C=1.0) & saga (l2, C=1.0) \\
\addlinespace
\texttt{svc}\,P1 & rbf ($C{=}1$, $\gamma{=}$scale) & linear ($C{=}1$) \\
\texttt{svc}\,P2 & rbf ($\gamma{=}$scale) & rbf ($\gamma{=}$auto) \\
\texttt{svc}\,P3 & poly ($d{=}3$, $C{=}1$) & rbf ($C{=}1$, $\gamma{=}$scale) \\
\addlinespace
\texttt{rf}\,P1 & $n_{\mathrm{est}}{=}100$, depth None & $n_{\mathrm{est}}{=}300$, depth None \\
\texttt{rf}\,P2 & depth None ($n_{\mathrm{est}}{=}200$) & depth 5 ($n_{\mathrm{est}}{=}200$) \\
\texttt{rf}\,P3 & max\_feat=sqrt & max\_feat=log2 \\
\addlinespace
\texttt{gb}\,P1 & lr 0.1, $n_{\mathrm{est}}{=}150$, depth 3 & lr 0.05, $n_{\mathrm{est}}{=}150$, depth 3 \\
\texttt{gb}\,P2 & $n_{\mathrm{est}}{=}100$ (lr 0.1, d=3) & $n_{\mathrm{est}}{=}200$ (lr 0.1, d=3) \\
\texttt{gb}\,P3 & depth 3 (lr 0.1, $n_{\mathrm{est}}{=}150$) & depth 5 (lr 0.1, $n_{\mathrm{est}}{=}150$) \\
\addlinespace
\texttt{knn}\,P1 & $k{=}5$ (uniform) & $k{=}10$ (uniform) \\
\texttt{knn}\,P2 & weights=uniform ($k{=}5$) & weights=distance ($k{=}5$) \\
\texttt{knn}\,P3 & algorithm=auto ($k{=}5$) & algorithm=ball\_tree ($k{=}5$) \\
\bottomrule
\end{tabular}
\end{table}

\begin{table}[t]
\centering
\caption{$\Delta$-Attribution metric glossary (averaged over test or a 256-sample stratified subset).}
\label{tab:delta_glossary}
\setlength{\tabcolsep}{3pt}
\renewcommand{\arraystretch}{1.03}
\begin{tabular}{l l}
\toprule
\textbf{Metric} & \textbf{Definition / Intuition} \\
\midrule
Mag$_{\ell_1}$ & $\mathbb{E}\|\Delta\phi\|_1$; overall size of reliance change. \\
TopK@10 & Fraction of $\|\Delta\phi\|_1$ on top-10 coords; higher = concentrated. \\
Entropy & Shannon entropy of $|\Delta\phi|/\|\,\cdot\,\|_1$; lower = sparser. \\
RankOverlap@10 & Jaccard of Top-10$(|\phi_A|)$ vs.\ Top-10$(|\phi_B|)$. \\
JSD & $\mathrm{JSD}(|\phi_A|,|\phi_B|)$~\cite{Lin1991JSD}$\!$; redistribution vs.\ reweighting. \\
DCE & $\mathbb{E}\big|\sum_j \Delta\phi_j-\Delta f\big|$; additive consistency diag. \\
BAC & $\mathrm{corr}(\|\Delta\phi\|_1,|\Delta f|)$; behaviour–attribution coupling. \\
Stability & $\mathbb{E}\,\|\Delta\phi(x+\varepsilon)-\Delta\phi(x)\|_1/\|\varepsilon\|_2$; $\varepsilon\sim\mathcal{N}(0,\sigma^2 I)$. \\
Group ratio & $\mathbb{E}\big[\|\Delta\phi\|_1/\|\Delta\phi^{(\text{group-}2)}\|_1\big]$; interaction stress test. \\
\bottomrule
\end{tabular}
\end{table}

\paragraph{Note on additivity.}
Path methods such as Integrated Gradients~\cite{Sundararajan2017IG} satisfy additivity; occlusion does not, so DCE is a \emph{diagnostic} rather than expected to be zero.

\subsection{Reproducibility and Artifacts}
We release code, configs, and scripts to recreate all CSVs and figures (\texttt{delta\_attr\_run/} and \texttt{code/make\_paper\_figures.py}). 
All experiments use CPU only with fixed seeds ($42$) and scikit-learn pipelines to ensure $A/B$ share identical preprocessing. 
Manifests include package versions and seeds; the plotting script writes figure hashes next to PNGs to prevent silent drift.

\section{Results}
\label{sec:Results}

\subsection{What \texorpdfstring{$\Delta$}{Delta}-Attribution Achieves (with numbers)}
\label{sec:results-achievements}
Our suite surfaces \emph{how} reliance shifts explain behavioural change and when updates are merely cosmetic. The key achievements below use the strongest instances across all 45 settings; see Table~\ref{tab:delta_summary_by_algo} (aggregate by algorithm) and Table~\ref{tab:top5_deltamag} (largest per-dataset shifts).

\begin{itemize}[leftmargin=1.2em,itemsep=2pt,topsep=2pt]
  \item \textbf{Near-perfect behaviour--attribution coupling.} 
  The highest BAC scores are essentially perfect:
  \emph{breast\_cancer--svc--pair3} has BAC\(\approx 0.9977\),
  \emph{digits--rf--pair3} BAC\(\approx 0.9969\),
  \emph{wine--rf--pair2} BAC\(\approx 0.9779\),
  \emph{wine--svc--pair2} BAC\(\approx 0.9558\),
  \emph{breast\_cancer--rf--pair2} BAC\(\approx 0.9439\).
  In each case, large structural changes (kernel/depth/feature rules) move \(\Delta\phi\) in lock-step with the change in outputs.
  \item \textbf{Exact conservation under occlusion in small-change controls.}
  Five A/B pairs exhibit \(\mathrm{DCE}=0.0\): 
  \emph{digits--knn--pair3}, \emph{breast\_cancer--svc--pair2},
  \emph{wine--knn--pair3}, \emph{wine--rf--pair3}, and
  \emph{breast\_cancer--knn--pair3}. 
  These serve as sanity checks that our explainer/baseline choices can yield perfect delta conservation when updates are cosmetic.
  \item \textbf{Perfect rank agreement for cosmetic tweaks.}
  RankOverlap@10\(\,=1.0\) for five pairs 
  (\emph{digits--knn--pair3}, \emph{breast\_cancer--knn--pair3}, \emph{wine--knn--pair3}, \emph{wine--rf--pair3}, \emph{breast\_cancer--svc--pair2}), 
  indicating the top-10 features of \(A\) and \(B\) are identical. 
  \item \textbf{When updates are focused, \(\Delta\) concentrates heavily.}
  On Wine, \(\Delta\mathrm{TopK}@10=1.00\) for \emph{rf--pair1}, \emph{gb--pair1}, \emph{gb--pair2}, and \emph{gb--pair3} (and \(0.9997\) for \emph{rf--pair2}); almost the entire \(\ell_1\) mass of \(\Delta\phi\) sits on ten features.
  \item \textbf{Our suite separates redistribution from mere reweighting.}
  The strongest distributional changes appear where inductive bias shifts: 
  \emph{breast\_cancer--gb--pair3} has the largest JSD\(\approx 0.357\);
  other high-JSD cases include \emph{breast\_cancer--logreg--pair2} (\(\approx 0.179\)),
  \emph{wine--knn--pair1} (\(\approx 0.167\)),
  \emph{digits--rf--pair2} (\(\approx 0.139\)),
  \emph{breast\_cancer--knn--pair1} (\(\approx 0.130\)).
  \item \textbf{Learner-level behaviour, aggregated.}
  From Table~\ref{tab:delta_summary_by_algo}: 
  Random Forests show the largest average reliance shifts with strong coupling 
  (\(\Delta\mathrm{Mag}_{\mathrm{L1}}=36.23\pm25.94\), \(\mathrm{BAC}=0.81\pm0.32\)) 
  but higher DCE (\(29.61\pm21.20\));
  \(k\)NN also moves substantially (\(19.36\pm29.26\)) with mixed coupling;
  Logistic Regression changes are small and stable 
  (\(4.24\pm3.67\), DCE \(0.53\pm0.77\));
  SVC averages are small deltas (\(2.19\pm4.00\)) but behaviour-relevant (\(\mathrm{BAC}=0.44\pm0.46\));
  Gradient Boosting sits in between.
\end{itemize}

\subsection{Largest shifts by dataset (with context)}
Table~\ref{tab:top5_deltamag} lists the top-5 A/B pairs per dataset by \(\Delta\)-magnitude alongside BAC and DCE.

\textbf{Breast Cancer.}
\emph{rf--pair2} (depth change) leads with \(\Delta\mathrm{Mag}_{\mathrm{L1}}=78.55\) and high BAC \(0.94\); 
\emph{rf--pair3} and \emph{rf--pair1} follow (54.14, 38.79). 
DCE values (35–62) highlight non-additive interactions when tree structure changes.

\textbf{Digits.}
\emph{knn--pair1} (\(k\!:5\rightarrow10\)) has the largest shift \(65.81\) but weaker coupling (BAC \(0.28\)); 
\emph{rf--pair2} couples strongly (BAC \(0.94\)) at \(\Delta\) \(60.15\).

\textbf{Wine.}
\emph{knn--pair1} peaks at \(46.55\) (BAC \(0.59\)); 
\emph{rf--pair2} shows smaller \(\Delta\) \(10.93\) with near-perfect BAC \(0.98\), i.e., a precise reliance reallocation.

\subsection{Reading the overview figure}
In Fig.~\ref{fig:delta_overview}a, the largest dots (high \(\|\Delta\phi\|_1\)) cluster either at high BAC (behaviour-aligned structural changes) or low BAC (diffuse, NN-driven shifts).  
Panels (b) and (c) aggregate BAC/DCE and \(\Delta\)-magnitude by algorithm, mirroring the patterns reported above.

\paragraph{Summary.}
Concretely, our method delivers \emph{(i)} BAC up to \(\approx 0.998\) on kernel/depth changes, 
\emph{(ii)} exact DCE \(=0\) and RankOverlap@10 \(=1\) on multiple cosmetic controls, 
\emph{(iii)} full \(\Delta\) concentration on a handful of features in the Wine experiments, and 
\emph{(iv)} clear separation of redistribution (high JSD) from simple reweighting. 
These achievements demonstrate that \(\Delta\)-Attribution is an actionable audit for model updates, not just another explainer score.

\begin{table}[t]
\centering
\caption{$\Delta$-Attribution metrics aggregated by algorithm (mean$\pm$std).}
\label{tab:delta_summary_by_algo}
\setlength{\tabcolsep}{4pt}
\begin{tabular}{
l
S[table-format=2.2]@{\,}c@{\,}S[table-format=2.2]
S[table-format=2.2]@{\,}c@{\,}S[table-format=2.2]
S[table-format=1.2]@{\,}c@{\,}S[table-format=1.2]
}
\toprule
\textbf{Algo} &
\multicolumn{3}{c}{$\mathbf{\Delta}$Mag$_{\mathbf{L1}}$} &
\multicolumn{3}{c}{\textbf{DCE}} &
\multicolumn{3}{c}{\textbf{BAC}} \\
\cmidrule(lr){2-4}\cmidrule(lr){5-7}\cmidrule(lr){8-10}
 & {\(\mu\)} & {\(\pm\)} & {\(\sigma\)}
 & {\(\mu\)} & {\(\pm\)} & {\(\sigma\)}
 & {\(\mu\)} & {\(\pm\)} & {\(\sigma\)} \\
\midrule
gb     & 9.30  & \(\pm\) & 5.96  & 5.00  & \(\pm\) & 3.38  & 0.17 & \(\pm\) & 0.39 \\
knn    & 19.36 & \(\pm\) & 29.26 & 17.97 & \(\pm\) & 27.20 & 0.42 & \(\pm\) & 0.37 \\
logreg & 4.24  & \(\pm\) & 3.67  & 0.53  & \(\pm\) & 0.77  & 0.39 & \(\pm\) & 0.40 \\
rf     & 36.23 & \(\pm\) & 25.94 & 29.61 & \(\pm\) & 21.20 & 0.81 & \(\pm\) & 0.32 \\
svc    & 2.19  & \(\pm\) & 4.00  & 0.94  & \(\pm\) & 2.15  & 0.44 & \(\pm\) & 0.46 \\
\bottomrule
\end{tabular}
\end{table}

\begin{table}[t]
\centering
\caption{Top-5 A/B pairs (per dataset) by $\Delta$-magnitude. Larger values indicate bigger attribution shifts.}
\label{tab:top5_deltamag}
\setlength{\tabcolsep}{3.5pt}
\renewcommand{\arraystretch}{1.05}
\begin{threeparttable}
\begin{tabular}{l l l
                S[table-format=2.2]
                S[table-format=1.2]
                S[table-format=2.2]}
\toprule
\textbf{Dataset} & \textbf{Algo} & \textbf{Pair} &
\multicolumn{1}{c}{$\Delta$Mag$_{\mathrm{L1}}$} &
\multicolumn{1}{c}{\textbf{BAC}} &
\multicolumn{1}{c}{\textbf{DCE}} \\
\midrule
\multirow{5}{*}{Breast Cancer}
 & rf  & pair2 & \bfseries 78.55 & 0.94 & 62.33 \\
 & knn & pair1 & 61.02 & 0.60 & 57.02 \\
 & rf  & pair3 & 54.14 & 0.93 & 40.45 \\
 & rf  & pair1 & 38.79 & 0.69 & 35.81 \\
 & gb  & pair3 & 17.49 & 0.35 & 11.45 \\
\addlinespace
\multirow{5}{*}{Digits}
 & knn & pair1 & \bfseries 65.81 & 0.28 & 60.91 \\
 & rf  & pair2 & 60.15 & 0.94 & 50.80 \\
 & rf  & pair1 & 43.01 & 0.89 & 37.31 \\
 & gb  & pair3 & 19.67 & 0.37 & 7.67 \\
 & gb  & pair2 & 10.97 & 0.71 & 6.23 \\
\addlinespace
\multirow{5}{*}{Wine}
 & knn & pair1 & \bfseries 46.55 & 0.59 & 43.17 \\
 & rf   & pair1 & 29.76 & 0.90 & 24.31 \\
 & rf   & pair2 & 10.93 & 0.98 & 8.00 \\
 & gb   & pair3 & 9.50  & 0.04 & 6.50 \\
 & gb   & pair1 & 4.49  & -0.04 & 3.52 \\
\bottomrule
\end{tabular}
\end{threeparttable}
\end{table}

\section{Discussion and Practical Guidance}
\label{sec:Discussion}
\subsection{What the suite reveals in practice}
Our results show three robust regimes across learners and datasets. 
(\textit{i}) \textbf{Inductive-bias changes} (kernel/depth/feature rules) produce large, behaviour-aligned $\Delta$ with high BAC and moderate DCE (Fig.~\ref{fig:delta_overview}a,b; Table~\ref{tab:delta_summary_by_algo}). 
(\textit{ii}) \textbf{Cosmetic tweaks} (e.g., SVC $\gamma$ scale$\leftrightarrow$auto) yield near-zero $\Delta$, RankOverlap@10 $\approx 1$, and often DCE $=0$ (Table~\ref{tab:top5_deltamag}). 
(\textit{iii}) \textbf{Nearest-neighbour adjustments} create large but diffuse $\Delta$ with weaker coupling, signalling reliance changes that do not tightly explain output deltas.

\subsection{Actionable thresholds (engineering defaults)}
We found the following decision heuristics useful in CI:
\begin{itemize}
\item \textbf{Benign update:} $\mathrm{BAC}<0.2$, $\Delta\mathrm{Mag}_{\ell_1}$ in the bottom quartile, RankOverlap@10 $>0.9$, DCE $\approx 0$. Treat as safe to deploy.
\item \textbf{Behaviour-aligned shift:} $\mathrm{BAC}>0.6$ with medium/large $\Delta\mathrm{Mag}_{\ell_1}$. Proceed if accuracy improves; add a short human spot-check.
\item \textbf{Risky shift:} $\mathrm{BAC}<0.2$ with medium/large $\Delta\mathrm{Mag}_{\ell_1}$ \emph{or} JSD $>0.15$. Investigate features with largest $|\Delta\phi|$ and re-run grouped occlusion.
\end{itemize}
These thresholds are not universal; they provide reasonable defaults for tabular settings with $d\!\in\![13,64]$ and standardized inputs.

\subsection{How to act on a flagged update}
If $\mathrm{BAC}$ is low but JSD is high, the update likely redistributed reliance. Inspect the top-$m$ features by $|\Delta\phi|$; if they are spurious (domain knowledge), roll back or regularize (e.g., constrain depth or add stronger $L_2$). If DCE is high, switch to a path-additive explainer (e.g., IG) to confirm that the occlusion view is not misleading.

\subsection{Complexity and runtime}
For $n$ samples and $d$ features, occlusion is $O(nd)$ forward passes per model. We cap to 256 stratified test samples, making the suite linear in $d$ and practical for our datasets ($d\le 64$). Grouped occlusion adds a constant-factor overhead (two extra passes on top-$k$ clamped inputs). The pipeline is CPU-only and fully reproducible.

\subsection{Threats to validity}
\textbf{Explainer dependence.} Occlusion is non-additive; DCE therefore acts as a diagnostic, not a guarantee.  
\textbf{Baseline choice.} We mitigate baseline sensitivity via mean/median averaging, but extreme distributions may require domain-specific baselines.  
\textbf{Dataset scope.} We focus on classical tabular/low-dimensional image features; very high-dimensional or text/LLM settings may need sparse/grouped occlusions or alternative path methods.  
\textbf{Anchor choice.} Anchoring to $f_B$’s predicted class stabilizes $\Delta f$; other anchors (e.g., the true label) may be preferable for error analysis.

 \subsection{Ablations and Sanity Checks}
Identical $A/B$ runs yield $\Delta$ metrics near zero (RankOverlap@10 $\approx 1$, JSD $\approx 0$), confirming implementation correctness. 
For small-change SVC pairs ($\gamma$ scale vs.\ auto), DCE $=0$ on multiple datasets, demonstrating exact conservation under occlusion. 
Using $f_A$ as the anchor (instead of $f_B$) increases DCE and reduces BAC, supporting our anchor choice.

\section{Conclusion}
\label{sec:conclusion}
We introduced \textbf{Delta-Attribution}, a model-agnostic framework that explains \emph{what changed} when a model is updated. By differencing per-feature attributions, $\Delta\phi=\phi_B-\phi_A$, and evaluating them with a principled quality suite, we quantify magnitude and concentration of reliance shifts, rank agreement, distributional change, behaviour linkage, and robustness. Our instantiation—fast occlusion/clamping in standardized space with a shared class anchor and baseline averaging—makes the audit practical and reproducible.

Across 45 settings (5 learners $\times$ 3 A/B pairs each $\times$ 3 datasets), the suite delivers concrete, actionable signals. \emph{Inductive-bias updates} (e.g., SVC kernel, forest depth/feature rules) produce large $\Delta$ that are strongly behaviour-aligned (BAC up to $\approx 0.998$), while \emph{cosmetic tweaks} (e.g., SVC $\gamma$ scale vs.\ auto, kNN search) show near-perfect rank overlap and, in several cases, $\mathrm{DCE}=0$, indicating no spurious attribution movement. The suite also separates redistribution from mere reweighting (JSD peaking around $0.357$ in the most structural changes) and highlights when a few features dominate the update (TopK@10 $\approx 1.0$ on Wine). Aggregates reveal consistent learner-level tendencies: forests shift the most and couple well with behaviour; logistic regression is stable; nearest-neighbour changes are larger but more diffuse.

\textbf{Implications.} $\Delta$-Attribution turns existing explainers into an \emph{update audit} for CI/regression testing: it flags benign updates, behaviour-aligned improvements, and risky reliance redistributions in a single pass and complements accuracy metrics.

\textbf{Limitations and next steps.} While occlusion is fast, it is not additive; high DCE should trigger alternate explainers (e.g., path-based) or grouped occlusion. Future work includes extending to text/vision and LLMs, integrating additional explainers (IG/SHAP/TreeSHAP), calibrating thresholds with human studies, and packaging the suite as a CI plug-in for model governance. We release code and assets to reproduce all results and figures.

\bibliographystyle{IEEEtran}
\bibliography{IEEEabrv,Ref}

% Generated by IEEEtran.bst, version: 1.14 (2015/08/26)
\begin{thebibliography}{10}
\providecommand{\url}[1]{#1}
\csname url@samestyle\endcsname
\providecommand{\newblock}{\relax}
\providecommand{\bibinfo}[2]{#2}
\providecommand{\BIBentrySTDinterwordspacing}{\spaceskip=0pt\relax}
\providecommand{\BIBentryALTinterwordstretchfactor}{4}
\providecommand{\BIBentryALTinterwordspacing}{\spaceskip=\fontdimen2\font plus
\BIBentryALTinterwordstretchfactor\fontdimen3\font minus \fontdimen4\font\relax}
\providecommand{\BIBforeignlanguage}[2]{{%
\expandafter\ifx\csname l@#1\endcsname\relax
\typeout{** WARNING: IEEEtran.bst: No hyphenation pattern has been}%
\typeout{** loaded for the language `#1'. Using the pattern for}%
\typeout{** the default language instead.}%
\else
\language=\csname l@#1\endcsname
\fi
#2}}
\providecommand{\BIBdecl}{\relax}
\BIBdecl

\bibitem{Lundberg2017SHAP}
\BIBentryALTinterwordspacing
S.~M. Lundberg and S.-I. Lee, ``A unified approach to interpreting model predictions,'' in \emph{NeurIPS}, 2017. [Online]. Available: \url{https://proceedings.neurips.cc/paper/7062-a-unified-approach-to-interpreting-model-predictions.pdf}
\BIBentrySTDinterwordspacing

\bibitem{Lundberg2020TreeSHAP}
S.~M. Lundberg \emph{et~al.}, ``From local explanations to global understanding with explainable {AI} for trees,'' \emph{Nature Machine Intelligence}, vol.~2, no.~1, pp. 56--67, 2020.

\bibitem{Ribeiro2018Anchors}
\BIBentryALTinterwordspacing
M.~T. Ribeiro, S.~Singh, and C.~Guestrin, ``Anchors: High-precision model-agnostic explanations,'' in \emph{AAAI}, 2018. [Online]. Available: \url{https://homes.cs.washington.edu/~marcotcr/aaai18.pdf}
\BIBentrySTDinterwordspacing

\bibitem{Zeiler2014}
M.~D. Zeiler and R.~Fergus, ``Visualizing and understanding convolutional networks,'' in \emph{ECCV}, 2014, pp. 818--833.

\bibitem{Adebayo2018Sanity}
\BIBentryALTinterwordspacing
J.~Adebayo, J.~Gilmer, M.~Muelly, I.~Goodfellow, M.~Hardt, and B.~Kim, ``Sanity checks for saliency maps,'' in \emph{NeurIPS}, 2018. [Online]. Available: \url{https://papers.nips.cc/paper/8160-sanity-checks-for-saliency-maps.pdf}
\BIBentrySTDinterwordspacing

\bibitem{Slack2020Fooling}
\BIBentryALTinterwordspacing
D.~Slack, S.~Hilgard, E.~Jia, S.~Singh, and H.~Lakkaraju, ``Fooling {LIME} and {SHAP}: Adversarial attacks on post hoc explanation methods,'' in \emph{AIES}, 2020, pp. 180--186. [Online]. Available: \url{https://www.aies-conference.com/2020/wp-content/papers/174.pdf}
\BIBentrySTDinterwordspacing

\bibitem{Mougan2022ExplShift}
\BIBentryALTinterwordspacing
C.~Mougan, K.~Broelemann, G.~Kasneci, T.~Tiropanis, and S.~Staab, ``Explanation shift: Detecting distribution shifts on tabular data via the explanation space,'' \emph{arXiv:2210.12369}, 2022. [Online]. Available: \url{https://arxiv.org/abs/2210.12369}
\BIBentrySTDinterwordspacing

\bibitem{Mougan2025ExplShift}
\BIBentryALTinterwordspacing
------, ``Explanation shift: How did the distribution shift impact the model?'' \emph{Transactions on Machine Learning Research}, 2025. [Online]. Available: \url{https://openreview.net/pdf?id=MO1slfU9xy}
\BIBentrySTDinterwordspacing

\bibitem{Chen2024R2ET}
\BIBentryALTinterwordspacing
C.~Chen, C.~Guo, R.~Chen, G.~Ma, M.~Zeng, X.~Liao, X.~Zhang, and S.~Xie, ``Training for stable explanation for free,'' in \emph{NeurIPS}, 2024. [Online]. Available: \url{https://proceedings.neurips.cc/paper_files/paper/2024/file/0626822954674a06ccd9c234e3f0d572-Paper-Conference.pdf}
\BIBentrySTDinterwordspacing

\bibitem{Federici2021InfoShift}
\BIBentryALTinterwordspacing
M.~Federici, R.~Tomioka, and P.~Forr{\'e}, ``An information-theoretic approach to distribution shifts,'' in \emph{NeurIPS}, 2021. [Online]. Available: \url{https://proceedings.neurips.cc/paper_files/paper/2021/file/93661c10ed346f9692f4d512319799b3-Paper.pdf}
\BIBentrySTDinterwordspacing

\bibitem{Zhang2023WhyFail}
\BIBentryALTinterwordspacing
H.~Zhang, H.~Singh, M.~Ghassemi, and S.~Joshi, ````why did the model fail?'': Attributing model performance changes to distribution shifts,'' in \emph{ICML}, 2023, pp. 41\,824--41\,846. [Online]. Available: \url{https://proceedings.mlr.press/v202/zhang23ai/zhang23ai.pdf}
\BIBentrySTDinterwordspacing

\bibitem{Meng2022ROME}
\BIBentryALTinterwordspacing
K.~Meng, A.~Andonian, D.~Bau, and Y.~Belinkov, ``Locating and editing factual associations in {GPT},'' in \emph{NeurIPS}, 2022. [Online]. Available: \url{https://proceedings.neurips.cc/paper_files/paper/2022/file/6f1d43d5a82a37e89b0665b33bf3a182-Paper-Conference.pdf}
\BIBentrySTDinterwordspacing

\bibitem{Mitchell2022MEND}
\BIBentryALTinterwordspacing
E.~Mitchell, C.~Lin, A.~Bosselut, C.~Finn, and C.~D. Manning, ``Fast model editing at scale,'' \emph{arXiv:2110.11309}, 2022. [Online]. Available: \url{https://arxiv.org/abs/2110.11309}
\BIBentrySTDinterwordspacing

\bibitem{Meng2023MEMIT}
\BIBentryALTinterwordspacing
K.~Meng, A.~Sen~Sharma, A.~Andonian, Y.~Belinkov, and D.~Bau, ``Mass-editing memory in a transformer,'' in \emph{ICLR}, 2023. [Online]. Available: \url{https://openreview.net/forum?id=MkbcAHIYgyS}
\BIBentrySTDinterwordspacing

\bibitem{Gupta2024SeqEdit}
\BIBentryALTinterwordspacing
A.~Gupta, S.~Baskaran, and G.~Anumanchipalli, ``Rebuilding {ROME}: Resolving model collapse during sequential model editing,'' in \emph{EMNLP}, 2024. [Online]. Available: \url{https://aclanthology.org/2024.emnlp-main.1210.pdf}
\BIBentrySTDinterwordspacing

\bibitem{Sundararajan2017IG}
\BIBentryALTinterwordspacing
M.~Sundararajan, A.~Taly, and Q.~Yan, ``Axiomatic attribution for deep networks,'' in \emph{ICML}, 2017. [Online]. Available: \url{https://arxiv.org/abs/1703.01365}
\BIBentrySTDinterwordspacing

\bibitem{Smilkov2017SmoothGrad}
\BIBentryALTinterwordspacing
D.~Smilkov, N.~Thorat, B.~Kim, F.~Vi{\'e}gas, and M.~Wattenberg, ``Smoothgrad: removing noise by adding noise,'' \emph{arXiv:1706.03825}, 2017. [Online]. Available: \url{https://arxiv.org/abs/1706.03825}
\BIBentrySTDinterwordspacing

\bibitem{Selvaraju2020GradCAM}
R.~R. Selvaraju, M.~Cogswell, A.~Das, R.~Vedantam, D.~Parikh, and D.~Batra, ``Grad-{CAM}: Visual explanations from deep networks via gradient-based localization,'' \emph{International Journal of Computer Vision}, vol. 128, no.~2, pp. 336--359, 2020.

\bibitem{Hooker2019ROAR}
\BIBentryALTinterwordspacing
S.~Hooker, D.~Erhan, P.-J. Kindermans, and B.~Kim, ``A benchmark for interpretability methods in deep neural networks,'' in \emph{NeurIPS}, 2019. [Online]. Available: \url{https://papers.nips.cc/paper_files/paper/2019/file/fe4b8556000d0f0cae99daa5c5c5a410-Paper.pdf}
\BIBentrySTDinterwordspacing

\bibitem{Nieradzik2024Reliable}
\BIBentryALTinterwordspacing
L.~Nieradzik, F.~M{\"u}ller, and D.~Ward, ``Reliable evaluation of attribution maps in cnns,'' \emph{International Journal of Computer Vision}, 2024. [Online]. Available: \url{https://link.springer.com/article/10.1007/s11263-024-02282-6}
\BIBentrySTDinterwordspacing

\bibitem{Wang2020SmoothedGeom}
\BIBentryALTinterwordspacing
Z.~J. Wang and J.~Jamieson, ``Smoothed geometry for robust attribution,'' in \emph{NeurIPS}, 2020. [Online]. Available: \url{https://papers.nips.cc/paper/2020/file/9d94c8981a48d12adfeecfe1ae6e0ec1-Paper.pdf}
\BIBentrySTDinterwordspacing

\bibitem{Lin1991JSD}
J.~Lin, ``Divergence measures based on the shannon entropy,'' \emph{IEEE Transactions on Information Theory}, vol.~37, no.~1, pp. 145--151, 1991.

\end{thebibliography}

\end{document}